\documentclass[conference]{IEEEtran}
\IEEEoverridecommandlockouts
\usepackage{cite}
\usepackage{amsmath,amssymb,amsfonts}
\usepackage{algorithmic}
\usepackage{graphicx}
\usepackage{textcomp}
\usepackage{xcolor}
\usepackage{multirow}
\usepackage{booktabs, bm}
\usepackage{subcaption, adjustbox}
\usepackage{enumitem}
\usepackage[hidelinks]{hyperref}
\usepackage{eso-pic}

\AddToShipoutPictureFG*{%
  \AtPageLowerLeft{%
    \raisebox{4em}{
      \makebox[\paperwidth]{\textit{Preprint}}%
    }%
  }%
}

\def\BibTeX{{\rm B\kern-.05em{\sc i\kern-.025em b}\kern-.08em
    T\kern-.1667em\lower.7ex\hbox{E}\kern-.125emX}}
\begin{document}

\title{Federated Diffusion Modeling with Differential Privacy for Tabular Data Synthesis}

\author{Timur Sattarov \thanks{Research conducted by Timur in part as a PhD candidate at the School of Computer Science, University of St.Gallen, Switzerland.} \\
Deutsche Bundesbank \\
Frankfurt am Main, Germany \\
timur.sattarov@bundesbank.de
\and
Marco Schreyer \thanks{Research conducted in part while Marco was with the International Computer Science Institute (ICSI), Berkeley, CA, USA.} \\
Swiss Federal Audit Office \\
Bern, Switzerland \\
marco.schreyer@efk.admin.ch
\and
Damian Borth \\
University of St.Gallen \\
St.Gallen, Switzerland \\
damian.borth@unisg.ch
}

\maketitle

\begin{abstract}
The increasing demand for privacy-preserving data analytics in various domains necessitates solutions for synthetic data generation that rigorously uphold privacy standards. We introduce the~\textit{DP-FedTabDiff} framework, a novel integration of \textit{Differential Privacy}, \textit{Federated Learning} and \textit{Denoising Diffusion Probabilistic Models} designed to generate high-fidelity synthetic tabular data. This framework ensures compliance with privacy regulations while maintaining data utility. We demonstrate the effectiveness of \textit{DP-FedTabDiff} on multiple real-world mixed-type tabular datasets, achieving significant improvements in privacy guarantees without compromising data quality. Our empirical evaluations reveal the optimal trade-offs between privacy budgets, client configurations, and federated optimization strategies. The results affirm the potential of \textit{DP-FedTabDiff} to enable secure data sharing and analytics in highly regulated domains, paving the way for further advances in federated learning and privacy-preserving data synthesis.

\end{abstract}


\section{Introduction}

The increasing demand for privacy-preserving data analytics in regulated industries, especially finance, has activated interest in federated and synthetic data solutions. Financial institutions, including central banks, supreme audit institutions, and commercial entities, collect detailed microdata to inform policy, assess credit risk, and detect fraud. However, such data is inherently sensitive and subject to strict privacy regulations such as the \textit{California Consumer Privacy Act} (CCPA)\footnote{\url{https://oag.ca.gov/privacy/ccpa}} or the European \textit{General Data Protection Regulation} (GDPR)\footnote{\url{https://eur-lex.europa.eu/eli/reg/2016/679/oj}}.

\vspace{0.1cm}

Despite these safeguards, the use of artificial intelligence (AI) and machine learning models in sensitive domains like finance or healthcare continues to be associated with significant privacy risks. Attacks such as membership inference or model inversion~\cite{fredrikson2015, salem2018} can expose personally identifiable or proprietary information~\cite{bender2021, kairouz2019, fredrikson2015, salem2018}. Moreover, data centralization is often prohibited or impractical due to regulatory, legal, or operational constraints.

\vspace{0.1cm}

\begin{figure}[t!]
  \centering
  \includegraphics[width=0.99\linewidth, keepaspectratio]{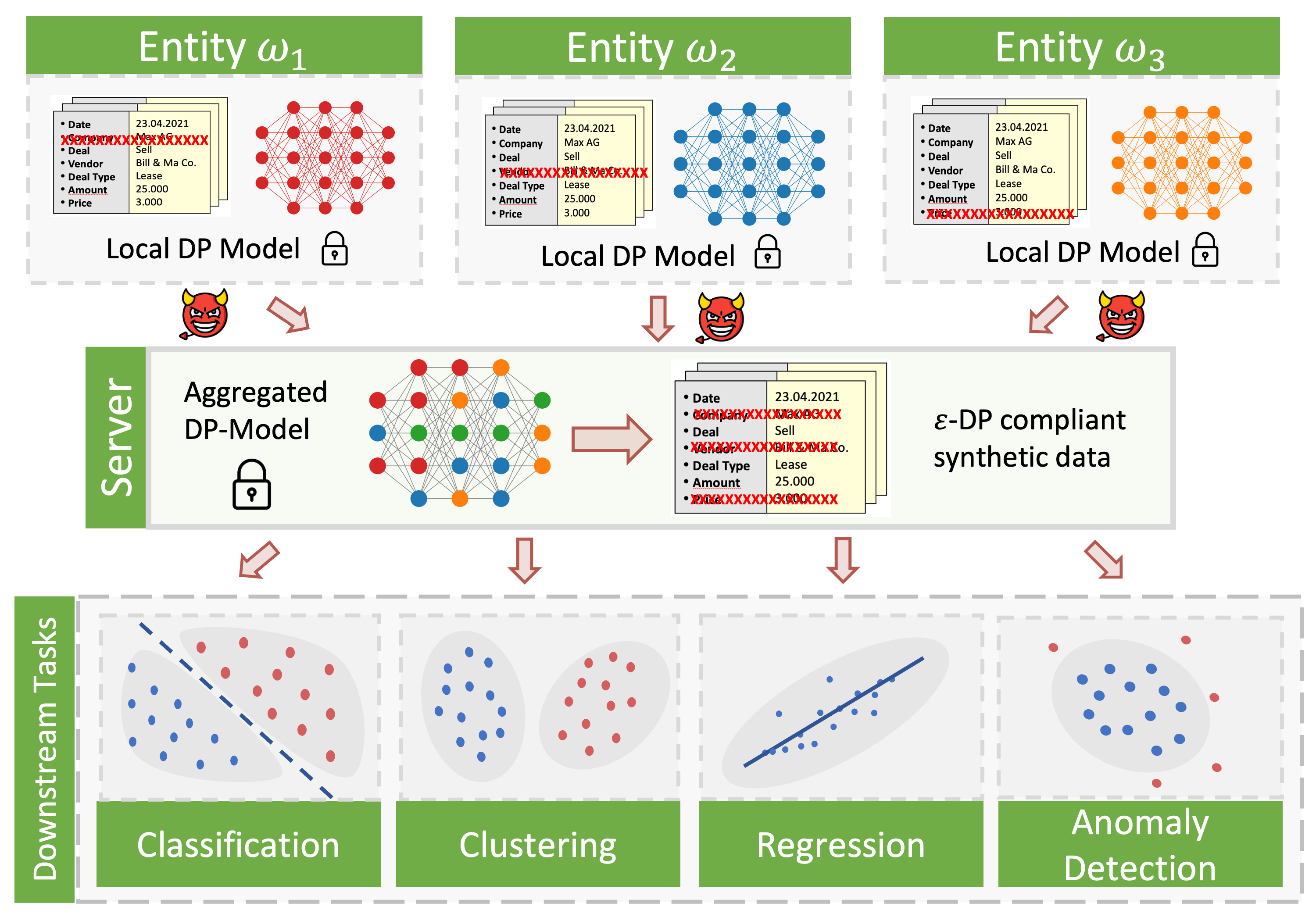}
    \vspace{-0.2cm}
    \caption{Overview of the proposed federated diffusion modeling with differential privacy (DP) for tabular data. Each entity trains a local DP diffusion model on sensitive data, and aggregation is protected by DP to mitigate model leakage. The aggregated model generates $\varepsilon$-DP compliant synthetic data for downstream tasks with formal privacy guarantees.}
  \label{fig:FinDiff_DP}
  \vspace{-0.3cm}
\end{figure}

This makes sensitive domains such as finance and healthcare a uniquely challenging and high-impact environment for \textit{Federated Learning} (FL). Here data is distributed across entities (i.e. hospitals, banks, jurisdictions), highly non-IID, often imbalanced, and governed by strict privacy obligations. Federated learning addresses these concerns by enabling decentralized model training without requiring raw data sharing~\cite{mcmahan2017a, mcmahan2017b}. However, FL alone does not guarantee resistance to inference attacks, as the shared model parameters can still leak sensitive information if intercepted or analyzed by an adversary (see Figure~\ref{fig:FinDiff_DP}, where the “devil” icons highlight such vulnerabilities). This risk motivates the integration of \textit{Differential Privacy} (DP), which mathematically bounds the influence of any individual data point on the learned model~\cite{dwork2006our}, thereby offering an additional layer of protection and significantly strengthening the privacy guarantees of FL systems.

\vspace{0.1cm}

Complementing these privacy-preserving mechanisms, synthetic data generation has emerged as a promising strategy to support safe data sharing and experimentation. High-fidelity synthetic tabular data can preserve analytical utility while avoiding direct exposure of real records. This is especially important in finance, where generating realistic but non-identifiable data is critical for risk modeling, fraud detection~\cite{charitou2021synthetic, Barse2003}, and simulation of rare events.


\begin{figure*}[t!]
  \centering
  \includegraphics[width=0.93\linewidth, ]{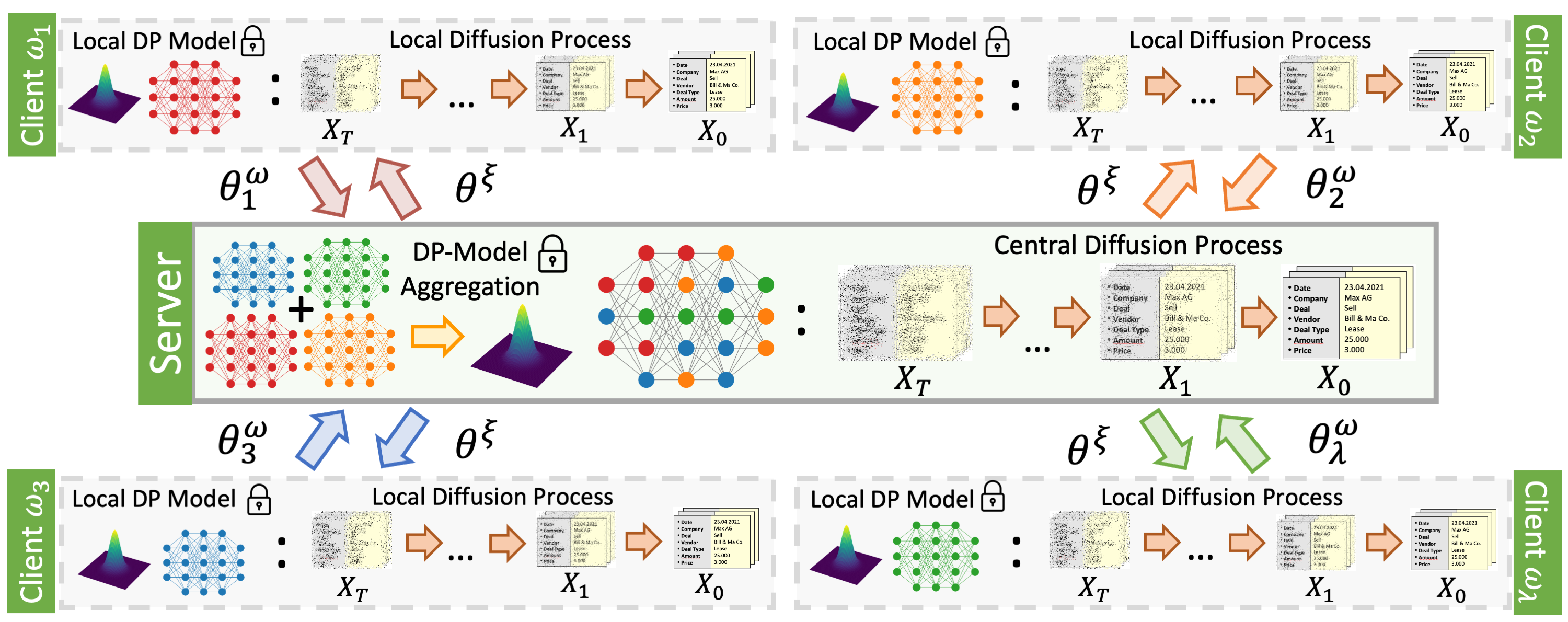}
    \caption{Schematic representation of the proposed \textit{DP-FedTabDiff} model. Each client $\omega_{i}$ independently trains a \textit{Federated Tabular Diffusion} (FedTabDiff)~\cite{sattarov2024fedtabdiff} model with \textit{Differential Privacy} (DP)~\cite{dwork2006our}. Timesteps $X_T, \ldots, X_0$ represent different stages of latent data representations in the generative reverse diffusion process. Client models $\theta_{i}^{\omega}$ are periodically aggregated into a global consolidated model $\theta^{\xi}$ on a central server, which is then redistributed back to each client for the next optimization round.}
    \vspace{-0.1cm}
  \label{fig:DP-FedTabDiff}
\end{figure*}

\vspace{0.1cm}

Recent progress in generative modeling has shown that \textit{Denoising Diffusion Probabilistic Models} (DDPMs) outperform GANs in stability, mode coverage, and sample quality, especially in complex, multimodal distributions~\cite{dhariwal2021diffusion, rombach2022high}. While most diffusion research has focused on image and text, recent work like \textit{FinDiff}~\cite{sattarov2023findiff} has demonstrated their viability for generating high-quality financial mixed-type tabular data. Building upon this, \textit{FedTabDiff}~\cite{sattarov2024fedtabdiff} extends the diffusion framework to the federated learning setting, allowing multiple institutions to train generative models collaboratively without centralizing sensitive data. However, neither approach offers formal privacy guarantees.

\vspace{0.1cm}

In this work, we propose a novel learning approach that integrates: (i) \textit{Differential Privacy}, (ii) \textit{Federated Learning}, and (iii) \textit{Denoising Diffusion Probabilistic Models} for synthesizing tabular data under regulatory constraints. Our framework is designed for environments where privacy, decentralization, and data heterogeneity are critical. The model builds upon \textit{FinDiff}~\cite{sattarov2023findiff} and \textit{FedTabDiff}~\cite{sattarov2024fedtabdiff} as foundational components. In summary, the main contributions we present are as follows:

\begin{itemize}
    \item The introduction of \textit{Differentially Private Federated Tabular Diffusion} framework (DP-FedTabDiff) to create synthetic tabular data with privacy protection guarantees.
    \item The framework allows for precise quantification and adjustment of the privacy budget to suit the unique confidentiality and privacy requirements.
    \item Comprehensive empirical evaluation using real-world mixed-type tabular datasets, demonstrating its effectiveness in synthesizing high-quality, privacy-compliant data.
\end{itemize}

\vspace{-0.1cm}

\section{Related Work}

Lately, diffusion models~\cite{cao2022survey, yang2022diffusion, croitoru2023diffusion} and federated learning~\cite{aledhari2020federated, li2021survey, zhang2021survey} have garnered significant research interest. The following literature review focuses on federated deep generative modeling of tabular data with differential privacy.

\vspace{0.1cm}

\noindent\textbf{Deep Generative Models:} Xu et al.~\cite{tvae_ctgan} introduced CTGAN, a conditional generator for tabular data, addressing mixed data types to surpass previous models' limitations. Building on GANs for oversampling, Engelmann and Lessmann~\cite{engelmann2021conditional} proposed a solution for class imbalances by integrating conditional Wasserstein GANs with auxiliary classifier loss. Jordon et al.~\cite{jordon2018pate} formulated PATE-GAN to enhance data synthesis privacy, providing DP guarantees by modifying the PATE framework. Torfi et al.~\cite{torfi2022differentially} presented a DP framework focusing on preserving synthetic healthcare data characteristics. Lin et al. introduced DoppelGANger, a GAN-based method for generating high-fidelity synthetic time series data \cite{lin2019generating}, and later analyzed the privacy risks of GAN-generated samples, highlighting vulnerabilities to membership inference attacks \cite{lin2021privacy}. To handle diverse data types more efficiently, Zhao et al.~\cite{zhao2021ctab} developed CTAB-GAN, a conditional table GAN that efficiently addresses data imbalance and distributions. 
Kotelnikov et al.\cite{kotelnikov2022tabddpm} explored tabular data modeling using multinomial diffusion models~\cite{hoogeboom2021argmax} and one-hot encodings, while \textit{FinDiff}~\cite{sattarov2023findiff}, foundational for our framework, uses embeddings for encoding. Recent models have emerged utilizing diffusion models to address the challenges of modeling tabular data, such as class imbalance \cite{schreyer2024imb, roy2024frauddiffuse}.

\vspace{0.1cm}

\noindent\textbf{Federated Deep Generative Models:} De Goede et al. in~\cite{de2023training} introduced a federated diffusion model framework utilizing Federated Averaging~\cite{mcmahan2017a} and a UNet backbone algorithm to train DDPMs on the Fashion-MNIST and CelebA datasets. This approach reduces the parameter exchange during training without compromising image quality. Concurrently, Jothiraj and Mashhadi in~\cite{jothiraj2023phoenix} introduce \textit{Phoenix}, an unconditional diffusion model that employs a UNet backbone to train DDPMs on the CIFAR-10 image database. Both studies underscore the pivotal role of federated learning techniques in advancing the domain. In the context of the mixed-type tabular data, Sattarov et al.~\cite{sattarov2024fedtabdiff} recently introduced \textit{FedTabDiff} model that merges federated learning with diffusion models.

\vspace{0.1cm}

\noindent\textbf{Differentially Private Federated Deep Generative Models}: 
The integration of \textit{Differential Privacy} (DP) proposed by Dwork et al.~\cite{dwork2006our}, into federated learning (FL) frameworks has gained considerable attention, particularly in enhancing the privacy of deep generative models~\cite{abadi2016deep, dockhorn2022differentially}. Fan et al. \cite{fan2020survey} provide a comprehensive survey of DP generative adversarial networks, emphasizing their potential in FL environments. Gargary and De Cristofaro \cite{gargary2024systematic} extend this by systematically reviewing federated generative models, including those leveraging DP. Specific implementations like Chen et al. \cite{chen2020gs}'s gradient-sanitized approach for DP GANs, and Lomurno et al. \cite{lomurno2022sgde}'s secure data exchange framework illustrate practical applications. Meanwhile, Augenstein et al. \cite{augenstein2019generative} discuss deep generative models in FL settings to maintain privacy across decentralized datasets. Additionally, Zhang et al. \cite{zhang2021feddpgan} demonstrate the use of federated DP GANs in detecting COVID-19 pneumonia, showcasing a critical healthcare application. In the financial domain, initial steps have been taken for applications such as fraud detection~\cite{byrd2020differentially}, or anomaly detection~\cite{schreyer2022federated}. 

\vspace{0.1cm}

These advances highlight the synergy between differential privacy and federated learning to develop privacy-preserving deep generative models. To the best of our knowledge, this is the first attempt using DP diffusion models in a federated learning setup for synthesizing mixed-type tabular data.

\section{Differentially Private Federated Diffusion}

This section details our proposed \textit{DP-FedTabDiff} model, which integrates Denoising Diffusion Probabilistic Models (DDPMs) with Federated Learning (FL) and enhances it with Differential Privacy (DP) for tabular data generation.

\vspace{0.1cm}

\noindent\textbf{Gaussian Diffusion Models.} The \textit{Denoising Diffusion Probabilistic Model}~\cite{sohl2015deep, ho2020denoising} operates as a latent variable model that incrementally perturbs data~\scalebox{0.95}{$x_0 \in \mathbb{R}^d$} with Gaussian noise~\scalebox{0.95}{$\epsilon$} through a forward process and restores it using a reverse process. Starting from~\scalebox{0.95}{$x_0$}, latent variables~\scalebox{0.90}{$x_1, \ldots, x_T$} are generated via a Markov Chain, transforming them into Gaussian noise~\scalebox{0.90}{$x_T \sim \mathcal{N}(0, I)$}, defined as:

\vspace{-0.1cm}

\begin{equation}
    q(x_t|x_{t-1})=\mathcal{N}(x_t;\sqrt{1-\beta_t}x_{t-1}, \beta_t I).
    \label{eq:q_step}
\end{equation}

\vspace{0.1cm}

\noindent In this context,~\scalebox{0.95}{$\beta_t$} represents the noise level at timestep~\scalebox{0.95}{$t$}. Sampling~\scalebox{0.95}{$x_t$} from~\scalebox{0.95}{$x_0$} is expressed as \scalebox{0.85}{$q(x_t|x_0)=\mathcal{N}(x_t;{\textstyle\sqrt{1-\hat{\beta_t}}} x_0, \hat{\beta_t} I)$}, where \scalebox{0.85}{$\hat{\beta_t}=1-\prod_{i=0}^{t} (1-\beta_i)$}. In the reverse process, the model denoises~\scalebox{0.95}{$x_t$} to recover~\scalebox{0.95}{$x_0$}. A neural network parameterized by~\scalebox{0.95}{$\theta$} is trained to approximate each step as \scalebox{0.95}{$p_\theta(x_{t-1}|x_t)=\mathcal{N}(x_{t-1}; \mu_\theta(x_t, t), \Sigma_\theta(x_t,t))$}, where~\scalebox{0.95}{$\mu_\theta$} and~\scalebox{0.95}{$\Sigma_\theta$} are the estimated mean and covariance. According to Ho et al.~\cite{ho2020denoising}, with~\scalebox{0.95}{$\Sigma_\theta$} being diagonal,~\scalebox{0.95}{$\mu_\theta$} is calculated as:

\vspace{-0.2cm}

\begin{equation}
    \mu_\theta(x_t, t)=\frac{1}{\sqrt{\alpha_t}}(x_t - \frac{\beta_t}{\sqrt{1-\hat{\alpha_t}}} \epsilon_\theta(x_t, t)).
\end{equation}


\noindent Here,~\scalebox{0.95}{$\alpha_t := 1-\beta_t$},~\scalebox{0.95}{$\hat{\alpha_t} := \prod_{i=0}^{t} \alpha_i$}, and~\scalebox{0.95}{$\epsilon_\theta(x_t, t)$} represents the predicted noise component. Empirical evidence suggests that using a simplified MSE loss yields better results compared to the variational lower bound~\scalebox{0.95}{$\log p_{\theta}(x_0)$}, as given by:

\vspace{-0.1cm}

\begin{equation}
    \mathcal{L}_t=\mathbb{E}_{x_0,\epsilon,t}||\epsilon-\epsilon_\theta(x_t,t)||_2^2.
\end{equation}

\noindent We employ \textit{FinDiff}~\cite{sattarov2023findiff} as the denoising diffusion probabilistic model designed for mixed-type tabular data modality.

\vspace{0.1cm}

\noindent\textbf{Federated Learning.} The training of DDPMs is enhanced through \textit{Federated Learning} (FL)~\cite{mcmahan2017a}, which enables learning from data distributed across multiple clients, denoted as~\scalebox{0.95}{$\{\omega_{i}\}^{\mathcal{C}}_{i=1}$}. The overall dataset is divided into subsets,~\scalebox{0.95}{$\mathcal{D}=\{\mathcal{D}_{i}\}^{\mathcal{C}}_{i=1}$}, each accessible by a single client~\scalebox{0.95}{$\omega_{i}$}, with varied data distributions. We employ \textit{FedTabDiff}~\cite{sattarov2024fedtabdiff}, an extension of \textit{FinDiff}~\cite{sattarov2023findiff}, in a federated setting. A central \textit{FinDiff} model~\scalebox{0.95}{$f^{\xi}_{\theta}$} with parameters~\scalebox{0.95}{$\theta^{\xi}$} is collaboratively learned by clients. Each client~\scalebox{0.95}{$\omega_{i}$} retains a decentralized \textit{FinDiff} model~\scalebox{0.95}{$f^{\omega}_{\theta, i}$} and contributes to the central model's training through synchronous updates across~\scalebox{0.95}{$r=1, \dots, \mathcal{R}$} communication rounds. A subset of clients~\scalebox{0.95}{$\omega_{i, r} \subseteq \{\omega_{i}\}^{\mathcal{C}}_{i=1}$} is selected each round, receiving the central model parameters~\scalebox{0.95}{$\theta^{\xi}_{r}$}, performing \scalebox{0.95}{$\gamma=1, \dots, \Gamma$} local optimization updates, and sending updated parameters back for aggregation. Fig. \ref{fig:DP-FedTabDiff} illustrates the entire process using four clients. A weighted average of the updates (i.e. \textit{Federated Averaging}~\cite{mcmahan2017a}) is used to compute the central model parameters, defined as:

\vspace{-0.2cm}

\begin{equation}
    \theta^{\xi}_{r+1} \leftarrow \frac{1}{|\mathcal{D}|} \sum_{i=1}^{\lambda} |\mathcal{D}_{i}| \; \theta^{\;\omega}_{i,r+1} \,,
    \label{equ::federated_learning::federated_averaging}
\end{equation}

\vspace{-0.1cm}

\noindent where~\scalebox{0.95}{$\lambda$} is the number of participating clients,~\scalebox{0.95}{$\theta^{\xi}_{r}$} the central, and~\scalebox{0.95}{$\theta^{\omega}_{i,r}$} the client model parameters,~\scalebox{0.95}{$r$} the communication round,~\scalebox{0.95}{$|\mathcal{D}|$} the total sample count, and~\scalebox{0.90}{$|\mathcal{D}_{i}| \subseteq |\mathcal{D}|$} the number of samples for client~\scalebox{0.95}{$\omega_i$}.

\vspace{0.1cm}

\noindent\textbf{Differential Privacy.} The concept of \textit{Differential Privacy} (DP)~\cite{dwork2006our} is a mathematical framework that ensures an algorithm's output does not significantly change when a single data point in the input is modified, protecting individual data points from inference. Formally, a randomized algorithm \( \mathcal{A} \) provides \((\varepsilon, \delta)\)-DP if for any two datasets \( D \) and \( D' \) differing by one element, and for any subset of outputs \( S \subseteq \text{Range}(\mathcal{A}) \):

\begin{equation}
    \mathbb{P}[\mathcal{A}(D) \in S] \leq e^\varepsilon \mathbb{P}[\mathcal{A}(D') \in S] + \delta,
\end{equation}

\noindent where \( \varepsilon \) is the privacy loss parameter (smaller \( \varepsilon \) means better privacy), and \( \delta \) is a small probability of failure.

\vspace{0.1cm}

\noindent\textbf{Federated Learning with Differential Privacy.} In the proposed \textit{DP-FedTabDiff} model, the parameter update process is modified to incorporate the \textit{Gaussian Mechanism}~\cite{dwork2014algorithmic}. For each minibatch, the gradient for each individual sample \( \nabla \ell(x_t, \theta) \) is computed and then clipped individually to a maximum norm \( C \). These clipped gradients are accumulated into a single gradient tensor, and Gaussian noise \( \mathcal{N}(0, \sigma^2 I) \) is added. The parameter \( \sigma \) is chosen based on the desired privacy budget \( \varepsilon \) and \( \delta \). Each client's local DP model update is computed as follows:

\begin{equation}
    \theta^{\omega}_{i,r+1} = \theta^{\omega}_{i,r} -\eta(\frac{1}{|B|} \sum_{x\in B} \text{clip}(\nabla \ell(x_t, \theta^{\omega}_{i,r}), C)  + \mathcal{N}(0, \sigma^2 I)),
\end{equation}

\noindent where $\eta$ denotes the learning rate and $B$ is the batch size. The central server aggregates these updates using the Federated Averaging technique defined in Equation~\eqref{equ::federated_learning::federated_averaging}, ensuring the privacy of individual client data (see \autoref{fig:DP-FedTabDiff}).

\section{Experimental Setup}

This section describes the details of the conducted experiments, encompassing datasets, data preparation steps, model architecture with hyperparameters, and evaluation metrics.

\subsection{Datasets and Data Preparation}
\label{subsec:datasets}

In our experiments, we utilized the following four real-world and mixed-type tabular datasets:

\begin{enumerate}[itemsep=4pt]

    \item \textbf{Credit Default}\footnote{\url{https://archive.ics.uci.edu/ml/datasets/default+of+credit+card+clients}} ($\mathcal{D}_{A}$): This dataset includes 30,000 customers default payments records (e.g., payment history and bill statements) from April to September 2005. Each record includes 9 categorical and 13 numerical attributes. 

    \item \textbf{Census Income}\footnote{\url{https://archive.ics.uci.edu/dataset/2/adult}} ($\mathcal{D}_{B}$): This dataset contains demographic information from the 1994 U.S. Census to predict whether a person earns more than \$50,000 per year. In total, there are 32,561 records each encompassing 10 categorical and 3 numerical attributes. 

    \item \textbf{Philadelphia City Payments}\footnote{\url{https://tinyurl.com/bdz2xdbx}} ($\mathcal{D}_{C}$): This dataset consists of 238,894 records. The payments were generated by 58 distinct city departments in 2017. Each payment includes 10 categorical and 1 numerical attribute(s). 

    \item \textbf{Marketing}\footnote{\url{https://tinyurl.com/zx4u8tf5}} ($\mathcal{D}_{D}$) This dataset contains 45,211 customer records of a bank from 2008 to 2010, used to predict whether a client will subscribe to a term deposit. Each record includes 10 categorical and 6 numerical attributes. 
    
\end{enumerate}

\begin{figure}
    \centering
    \includegraphics[width=0.99\linewidth]{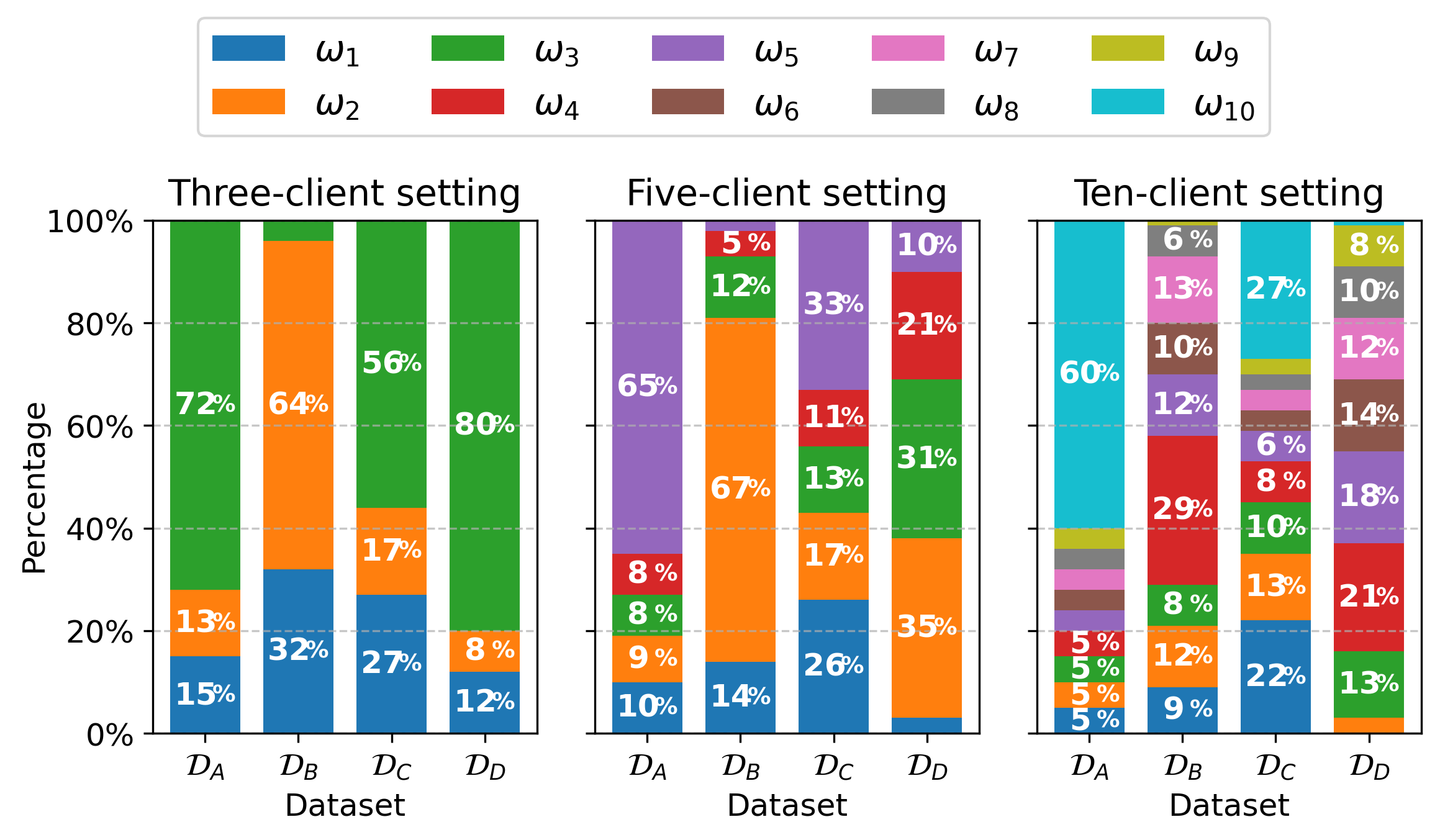}
    \caption{Non-IID data distribution among  (3, 5, and 10) client settings. Each bar indicates the percentage of data allocated to each client $\omega_i$ across four datasets.}
    \label{fig:noniid_splits}
\end{figure}

To simulate a realistic non-IID and unbalanced data environment for federated training, each dataset is partitioned based on a categorical feature. The descriptive statistics on the non-IID data partitioning schemes of 3, 5, and 10 client settings are presented in \autoref{fig:noniid_splits}. To standardize the numeric attributes, we employed quantile transformations\footnote{\url{https://tinyurl.com/ht9pz8m5}}. For embedding the categorical attributes, we followed the same approach as outlined in~\cite{sattarov2023findiff}.

\subsection{Model Architecture and Hyperparameters}

Next, we detail the architecture and hyperparameters chosen in \textit{DP-FedTabDiff} model optimization.

\vspace{0.1cm}

\noindent \textbf{Diffusion Model.\footnote{Model parameter optimization is conducted using \textit{PyTorch} v2.2.1~\cite{pytorch}.}} The architecture for all datasets consists of three layers, each comprising 1024 neurons, except for $\mathcal{D}_{C}$, which contains 2048 neurons. The models are trained for up to \scalebox{0.95}{$R=3,000$} communication rounds utilizing a mini-batch size of 16. 
The Adam optimizer~\cite{kingma2014adam} is utilized with parameters \scalebox{0.95}{$\beta_{1}=0.9$} and \scalebox{0.95}{$\beta_{2}=0.999$}. The hyperparameters of the underlying diffusion model \textit{FinDiff} are adopted from~\cite{sattarov2023findiff}. These settings include 500 diffusion steps (\scalebox{0.95}{$T=500$}) and a linear learning-rate scheduler with initial and final rates of \scalebox{0.95}{$\beta_{start}=0.0001$} and \scalebox{0.95}{$\beta_{end}=0.02$}, respectively. Each categorical attribute is represented as a 2-dimensional embedding.

\vspace{0.1cm}

\noindent \textbf{Federated Learning.\footnote{Federated learning is simulated using the \textit{Flower} framework v1.7.0~\cite{beutel2022flower}.}} In each communication round \scalebox{0.95}{\( r = 1, \ldots, R \)}, a random client \scalebox{0.95}{\( \omega_i \)} performs \scalebox{0.95}{\( \gamma=1, \dots, \Gamma \)} local optimization updates on its model \scalebox{0.95}{\( \theta^\omega_i \)} before sharing the updated parameters. The number of client optimization updates is evaluated across various settings \scalebox{0.95}{\( \Gamma \in [10, 50, 100, 500, 1000] \)}. Configurations with different numbers of clients \scalebox{0.95}{\( \lambda \in [3, 5, 10] \)} are also examined. Four distinct federated optimization strategies are explored: \textit{Federated Averaging} (\textit{FedAvg})~\cite{mcmahan2017a}, \textit{Federated Adam} (\textit{FedAdam})~\cite{reddi2020adaptive}, \textit{Federated Proximal} (\textit{FedProx})~\cite{li2020federated}, and \textit{Federated Yogi} (\textit{FedYogi})~\cite{reddi2020adaptive}. FedAvg aggregates client models by computing the weighted average of their parameters. FedAdam, an extension of the Adam optimizer, utilizes default hyperparameters \scalebox{0.95}{\( \beta_1 = 0.9 \)}, \scalebox{0.95}{\( \beta_2 = 0.999 \)}, and \scalebox{0.95}{\( \epsilon = 1e-8 \)}. FedProx introduces a proximal term to address client heterogeneity, employing a default \scalebox{0.95}{\( \mu = 0.01 \)}. Lastly, FedYogi adapts the Yogi optimizer using default parameters \scalebox{0.95}{\( \beta_1 = 0.9 \)}, \scalebox{0.95}{\( \beta_2 = 0.999 \)}, and \scalebox{0.95}{\( \epsilon = 1e-8 \)}.

\vspace{0.1cm}

\noindent \textbf{Differential Privacy.\footnote{Training with differential privacy is performed using \textit{Opacus} v1.4.1~\cite{opacus}.}}
We adopt the privacy settings from \cite{dockhorn2022differentially}, training models with $\varepsilon \in \{0.2, 1, 10\}$, corresponding to high, moderate, and low privacy levels. The probability of information leakage is set to the reciprocal of the number of training samples, $\delta = N^{-1}$, a common heuristic in practice.

\begin{figure*}[t!]
  \centering
  \begin{subfigure}[b]{0.49\linewidth}
    \centering
    \includegraphics[width=\linewidth]{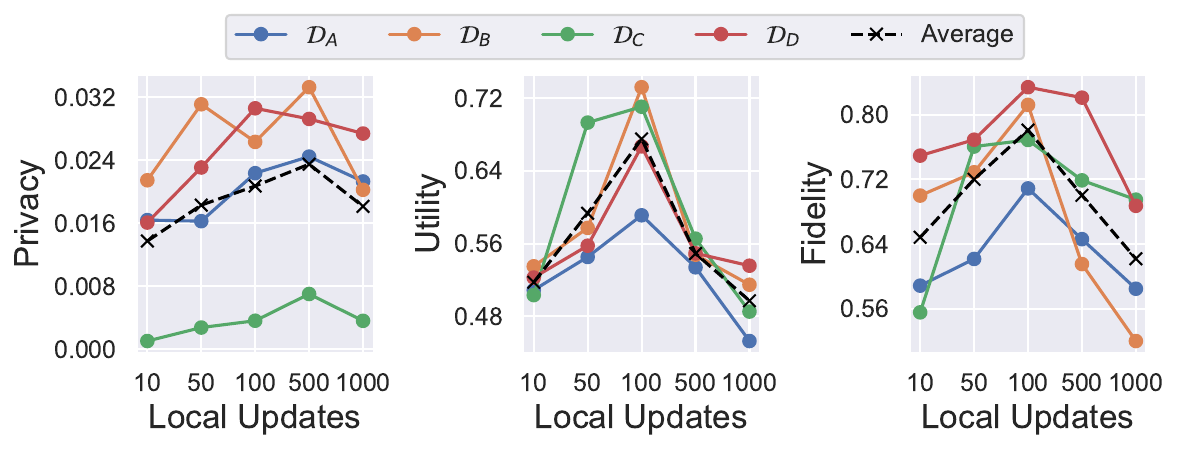}
    \caption{Non-IID Results}
    \label{fig:results_local_rounds_DP:noniid}
  \end{subfigure}
   \hfill
  \begin{subfigure}[b]{0.49\linewidth}
    \centering
    \includegraphics[width=\linewidth]{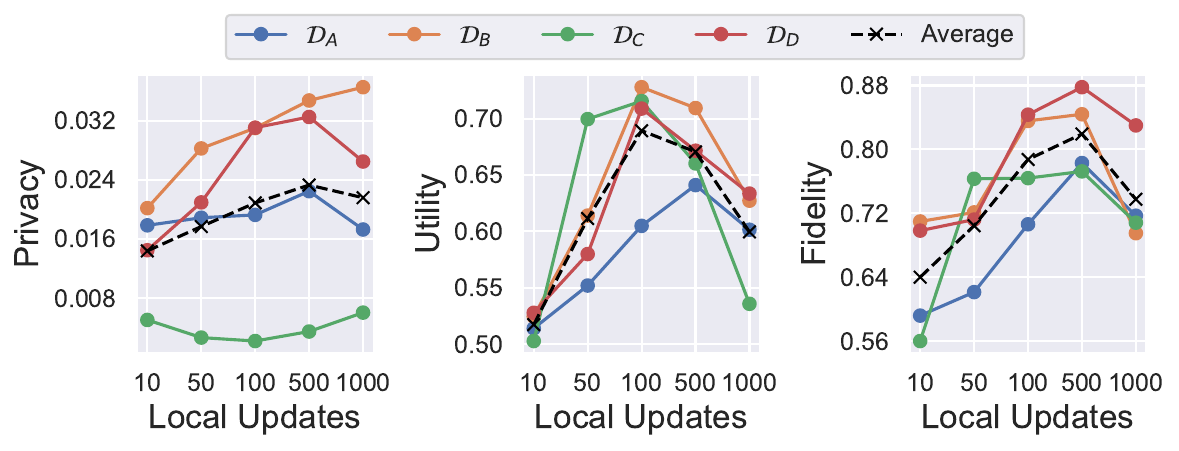}
    \caption{IID Results}
    \label{fig:results_local_rounds_DP:iid}
  \end{subfigure}
  \caption{Comparative evaluation of local optimization updates \scalebox{0.95}{$\Gamma \in [10, 50, 100, 500, 1000]$} between (a) non-IID and (b) IID settings, highlighting their impact on privacy, utility, and fidelity. In IID settings, increasing the number of local updates consistently improves performance, whereas, in non-IID settings, performance degrades after reaching a certain threshold.}  \label{fig:results_local_rounds_DP}
\end{figure*}

\begin{figure}
    \centering
    \includegraphics[width=1\linewidth]{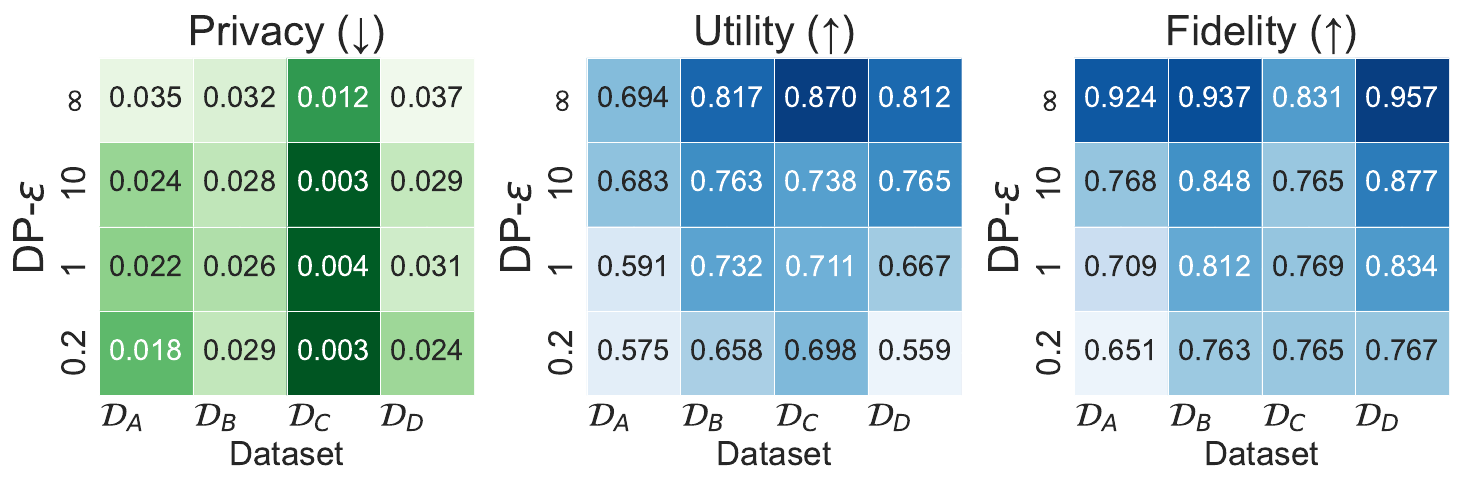}
    \caption{Heatmaps illustrating the impact of DP budgets $\varepsilon \in [0.2, 1, 10]$ and $\infty$ (no DP) on various datasets across Privacy, Utility, and Fidelity in non-IID settings. It is observed that as the DP budget decreases (i.e. lower $\varepsilon$ values), privacy protection improves, while fidelity and utility decline.}
  \label{fig:DP_results}
\end{figure}

\begin{figure*}[t!]
  \centering
  \begin{subfigure}[b]{0.19\linewidth}
    \centering
    \includegraphics[width=\linewidth, ]{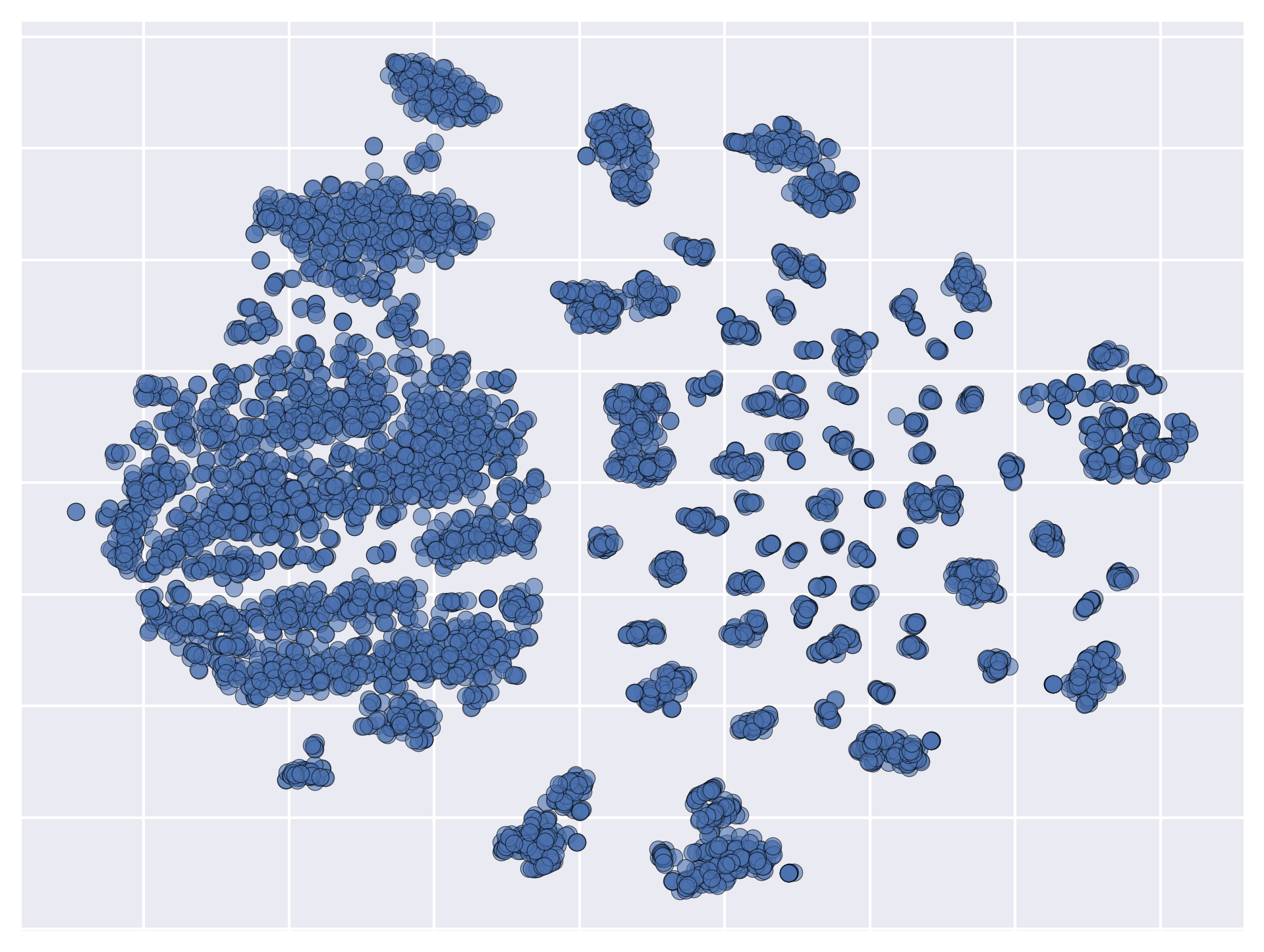}
    \caption{Original}
    \label{fig:tsne_real}
  \end{subfigure}
  \hfill
  \begin{subfigure}[b]{0.19\linewidth}
    \centering
    \includegraphics[width=\linewidth, ]{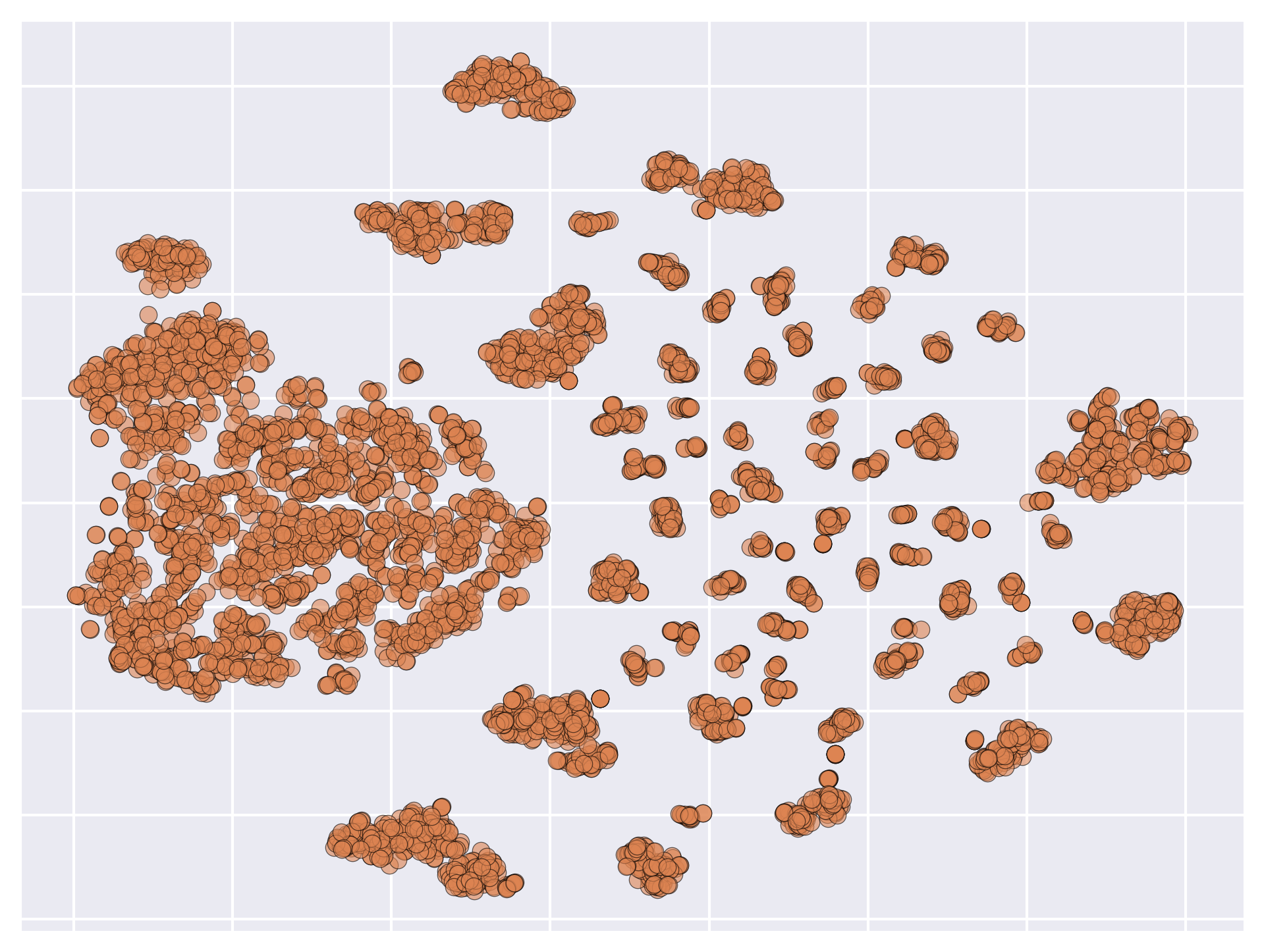}
    \caption{Synthetic ($\varepsilon$=$\infty$)}
    \label{fig:tsne_noDP}
  \end{subfigure}
    \hfill
  \begin{subfigure}[b]{0.19\linewidth}
    \centering
    \includegraphics[width=\linewidth, ]{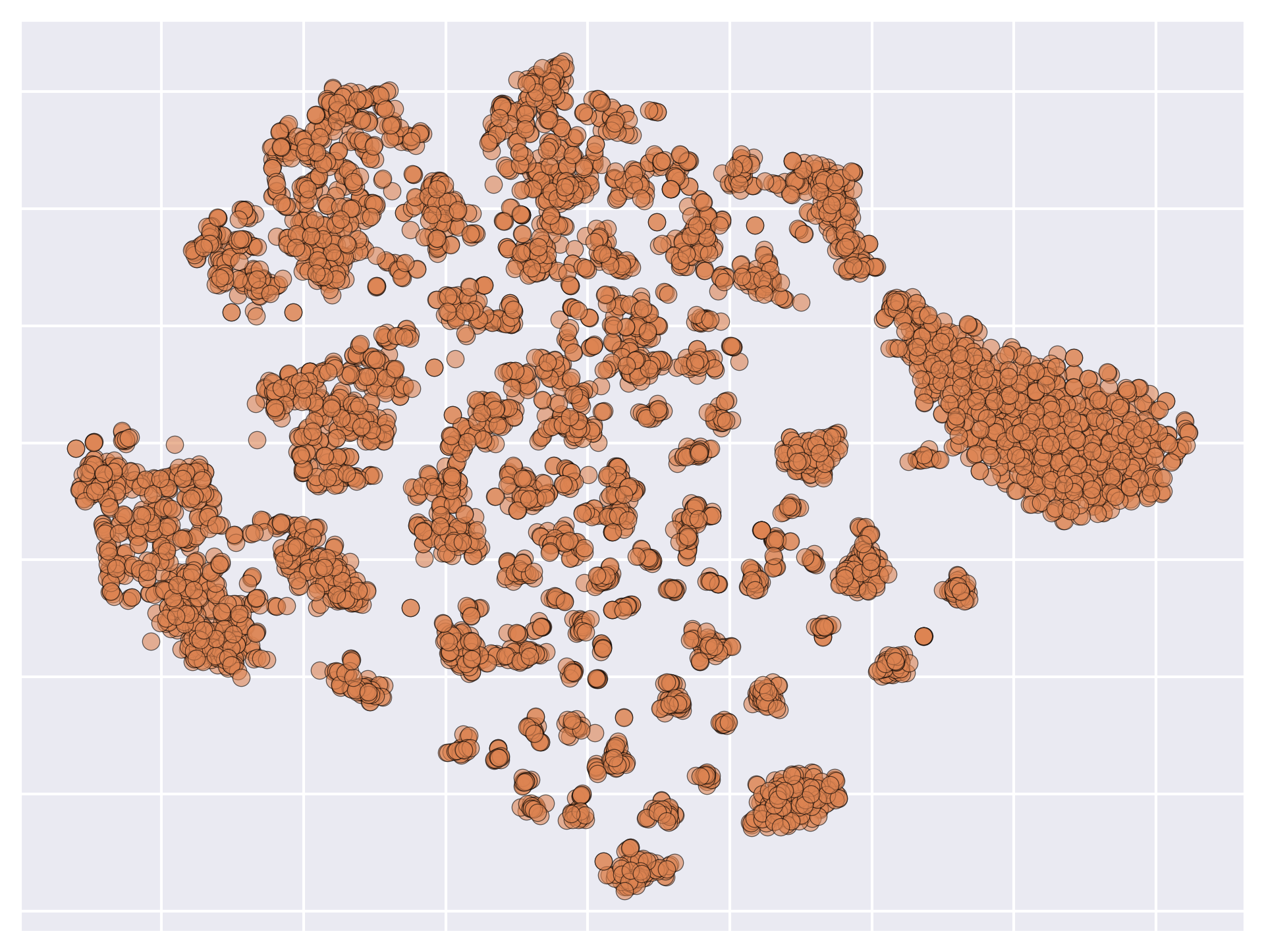}
    \caption{Synthetic ($\varepsilon$=10)}
    \label{fig:tsne_DP10}
  \end{subfigure}
    \hfill
  \begin{subfigure}[b]{0.19\linewidth}
    \centering
    \includegraphics[width=\linewidth, ]{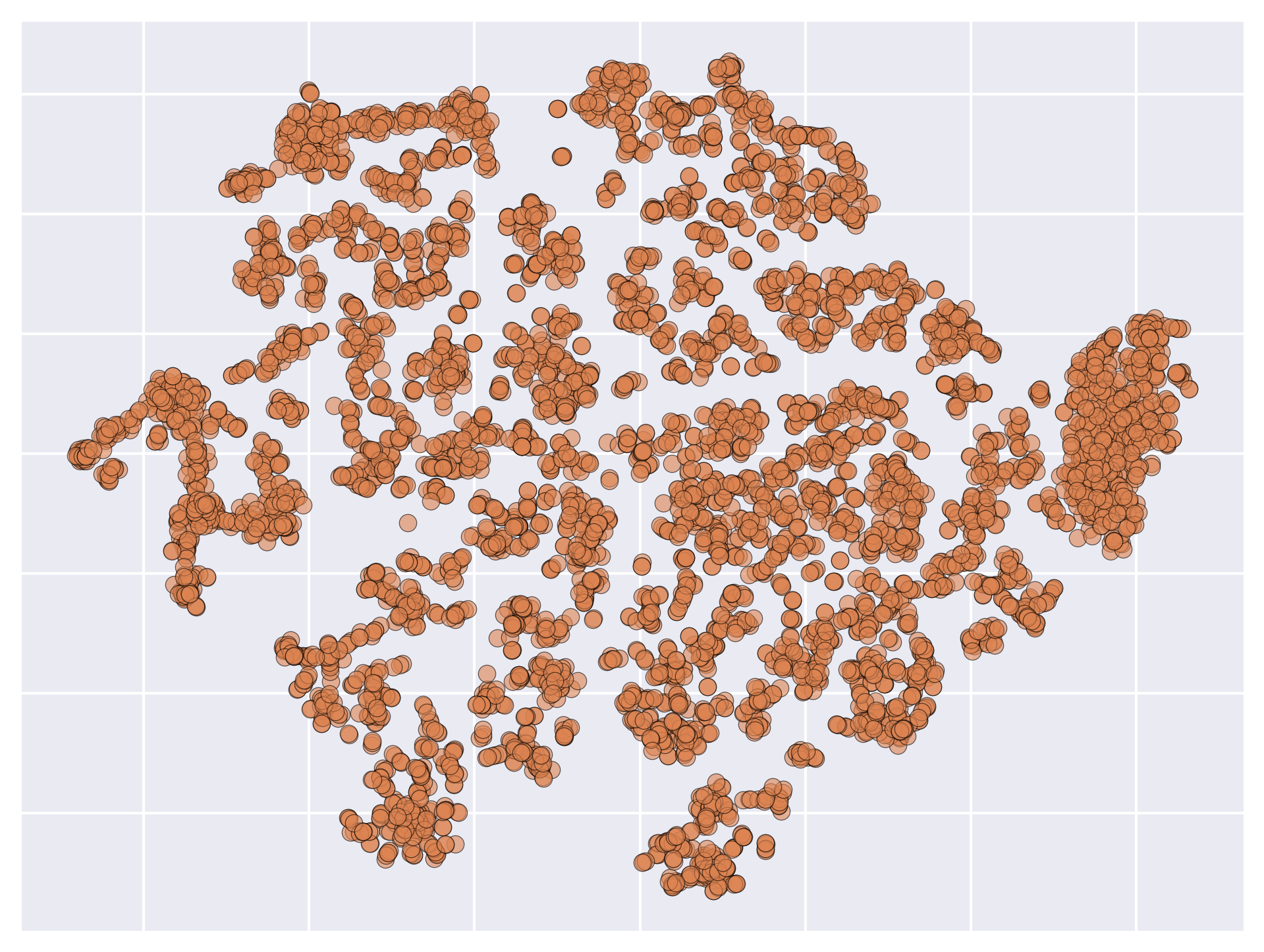}
    \caption{Synthetic ($\varepsilon$=1)}
    \label{fig:tsne_DP1}
  \end{subfigure}
    \hfill
  \begin{subfigure}[b]{0.19\linewidth}
    \centering
    \includegraphics[width=\linewidth, ]{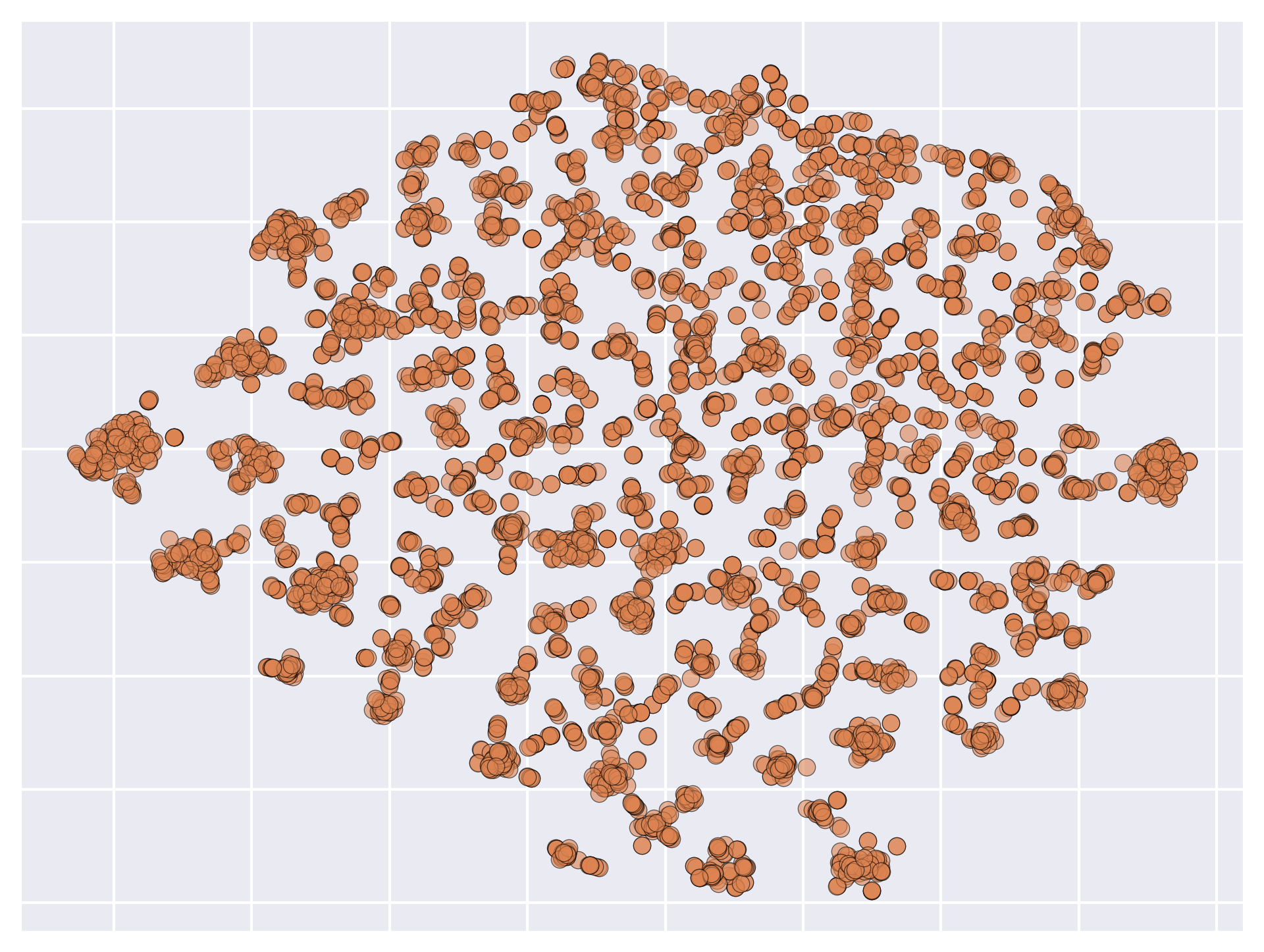}
    \caption{Synthetic ($\varepsilon$=0.2)}
    \label{fig:tsne_DP02}
  \end{subfigure}
  \caption{t-SNE visualization of (a) dataset $\mathcal{D}_{A}$, (b) synthetic data without DP, and (c-e) with DP using federated optimization across 5 clients ($\lambda$=5). As privacy increases (i.e., lower $\varepsilon$), DP noise degrades the structural integrity of the synthetic data.}
  \label{fig:TSNE}
\end{figure*}

\subsection{Evaluation Metrics}

A comprehensive set of evaluation metrics, including \textbf{privacy}, \textbf{utility}, and \textbf{fidelity}, is used to assess the model's effectiveness, offering a holistic view of data generation quality. 

\vspace{0.1cm}

\noindent \textbf{Privacy.}\footnote{The privacy risks estimation was conducted using the \textit{Anonymeter}~\cite{anonymeter}.} The privacy metric quantifies the extent to which synthetic data prevents the identification of original data entries. In this study, privacy is assessed using three key indicators of factual anonymization as outlined by the GDPR~\cite{article29wp216}. 
Specifically, we employ privacy evaluators to measure the risks of (i) \textit{singling out}, (ii) \textit{linkability}, and (iii) \textit{inference} that could potentially affect data donors following the release of synthetic datasets. These risk measures are defined as follows, with $\bm{S}$ for synthetic dataset and $\bm{X}$ for real dataset:

\begin{itemize}

\item \textit{Singling Out Risk} quantifies the probability  \( SOR(X, S) \) that a synthetic record uniquely corresponds to a real record. This risk assesses the likelihood that an individual in the real data can be identified based on a unique synthetic entry. Results include a 95\% confidence interval.

\item \textit{Linkability Risk} measures the proportion of successful attribute linkages   \( LR(X, S) \)  between synthetic and real records. It reflects the potential to link synthetic to real records via shared attributes, using a subset of 10 attributes (6 for the dataset $\mathcal{D_{C}}$).

\item \textit{Inference Risk} evaluates an attacker’s ability to predict a secret attribute using auxiliary data, quantified by model accuracy \( IR(X, S) \). This risk measures how well an adversary can infer unknown information, with each column as a secret and others as auxiliary data.

\end{itemize}

\noindent Each evaluator performs 500 attacks per record, computing the mean risk over all synthetic samples. The final privacy score aggregates results from all three evaluators, as defined by:

\vspace{-0.1cm}

\begin{equation}
    \Pi = \frac{1}{3} (SOR(X, S) + LR(X, S) + IR(X, S)).
\end{equation}

\noindent This empirical, attack-based evaluation framework ensures a robust assessment of privacy in synthetic data, reflecting real-world privacy risks more accurately than traditional metrics.

\vspace{0.1cm}

\noindent \textbf{Utility.} The effectiveness of synthetic data is determined by its utility, a measure of how functionally equivalent it is to real-world data. We quantify it by training classifiers on synthetic data ($S_{\text{train}}$), aligned dimensionally with the real training set, and evaluating on the real test set ($X_{\text{test}}$). This assesses how well statistical properties are preserved for model training. The average classifier accuracy represents overall utility, formalized as:

\vspace{-0.1cm}

\begin{equation}
\Phi = \frac{1}{N} \sum_{i=1}^{N} \Theta_i(S_{\text{train}}, X_{\text{test}}).
\end{equation}


\noindent Here, $\Phi$ represents the utility score, and $\Theta_i$ denotes the accuracy of the $i$-th classifier. To provide a comprehensive evaluation, we selected $N$=5 classifiers for this study, namely \textit{Random Forest}, \textit{Decision Trees}, \textit{Logistic Regression}, \textit{Ada Boost}, and \textit{MLP Classifier}.

\vspace{0.1cm}

\noindent \textbf{Fidelity.}\footnote{The row fidelity was computed using \textit{Dython} library v0.7.5~\cite{Zychlinski_dython_2018}.} Fidelity assesses how closely synthetic data emulates real data, considering both column-level and row-level comparisons. For column fidelity, the similarity between corresponding columns in synthetic and real datasets is evaluated. To measure the distance between the distributions of numeric attributes the \textit{Wasserstein similarity} \scalebox{0.95}{$WS(x^d, s^d)$} is employed. The \textit{Jensen-Shannon divergence}, denoted as \scalebox{0.95}{$JS(x^d, s^d)$} quantifies differences in categorical attributes. 
These metrics were combined to form the column fidelity score \scalebox{0.95}{$\Omega_{col}$} as:


\begin{equation}
    \Omega_{col}=\begin{cases}
        1-WS(x^d, s^d) & \text{if $d$ is num.} \\
        1-JS(x^d, s^d) & \text{if $d$ is cat.} 
    \end{cases}
\end{equation}

\noindent The overall fidelity for columns in synthetic dataset $S$ is the mean of \scalebox{0.95}{$\Omega_{col}(x^d, s^d)$} across all attributes. Row fidelity focuses on correlations between column pairs. For numeric attributes, the \textit{Pearson Correlation} between pairs \scalebox{0.95}{$\rho(x^a, x^b)$} is used. The discrepancy in correlations for real and synthetic pairs, \scalebox{0.95}{$PC(x^{a,b}, s^{a,b}) = |\rho(x^a, x^b) - \rho(s^a, s^b)|$}, quantifies this aspect. The \textit{Theil~U} coefficient quantifies the association between two categorical variables, denoted as \scalebox{0.95}{$TU(x^{a,b}, s^{a,b}$)}. 


\begin{equation}
    \Omega_{row}=\begin{cases}
        1-PC(x^{a,b}, s^{a,b}) & \text{if $d$ is num.} \\
        1-TU(x^{a,b}, s^{a,b}) & \text{if $d$ is cat.} 
    \end{cases}
\end{equation}

\noindent The total row fidelity for the dataset $S$ is the average of \scalebox{0.90}{$\Omega_{row}(x^{a,b}, s^{a,b})$} across all attribute pairs. Finally, the aggregate fidelity score \scalebox{0.90}{$\Omega(X, S)$} is the mean of column and row fidelity.

\vspace{0.1cm}

\section{Experimental results.}

This section presents the results of the experiments, demonstrating the efficacy of the \textit{DP-FedTabDiff} model and providing quantitative analyses. The conducted experiments are accompanied by three Research Questions (RQ).

\vspace{0.1cm}

\begin{figure}[t!]
    \centering
    \includegraphics[width=\linewidth]{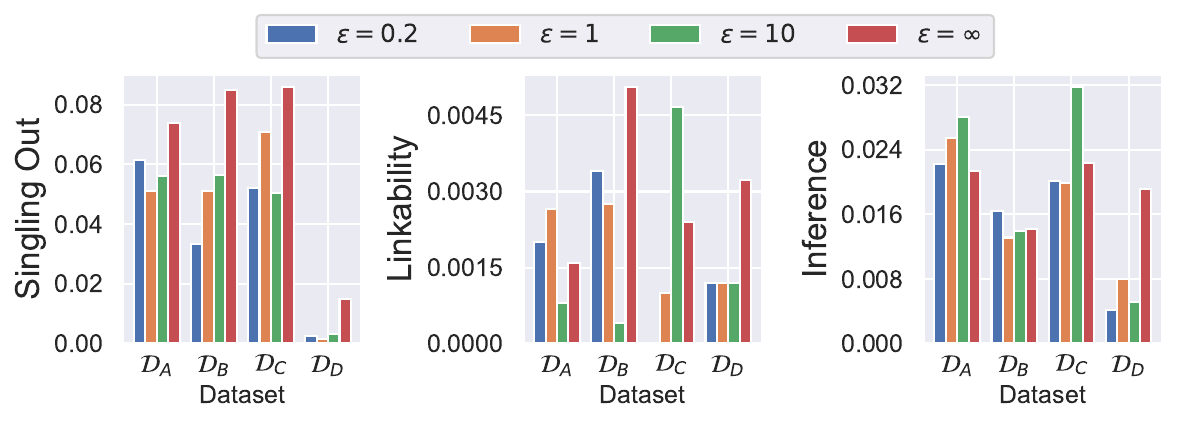}
    \caption{Singling Out, Linkability, and Inference risk evaluation under various privacy budgets ($\varepsilon$=$\infty$ denotes no DP) across all datasets ($\mathcal{D}_{A}$-$\mathcal{D}_{D}$). It is observed that the privacy risk metrics increase as the privacy budget increases.}  
    \label{fig:results_privacy_scores}
\end{figure}

\noindent \textbf{RQ 1:} \textit{How does varying number of local optimization updates impact the training of diffusion models for tabular data in a federated learning setup employing differential privacy?}

\vspace{0.1cm}

\noindent Minimizing model exchange frequency is crucial for privacy while maintaining the global model's generalization capabilities. We define \(\Gamma\) as the local updates performed by a client \(\omega_i\) at each round \(r\) before synchronization. We selected \(\Gamma\) values from \scalebox{0.95}{\([10, 50, 100, 500, 1000]\)} and conducted experiments in both IID and non-IID settings, fixing the privacy budget at \(\varepsilon=1\) with five federated clients (\(\lambda=5\)).

\vspace{0.1cm}

\noindent \textbf{Results:} We observed an increase in the average privacy risk across all datasets with more local optimization updates in both IID and non-IID settings, peaking at 500 updates (see \autoref{fig:results_local_rounds_DP}). The results suggest that more local updates reduce privacy protection.

\vspace{0.1cm}

For utility and fidelity, increasing the number of client updates in the IID setting (see \autoref{fig:results_local_rounds_DP:iid}) consistently improved performance. Each client received an IID data partition, which helped maintain model quality. However, in the non-IID setting (see \autoref{fig:results_local_rounds_DP:noniid}), performance declined after a threshold of \(\Gamma = 100\). The decline is caused by heterogeneous data distributions among clients. As a result, the clients drift away from a globally optimal model \cite{karimireddy2019scaffold}, leading to unstable and slow convergence.

\vspace{0.1cm}

Our findings indicate that 100 local updates provide an optimal trade-off between privacy, utility, and fidelity in both IID and non-IID settings. Additionally, this choice significantly reduced the overall training time from 28 hours (\(\Gamma = 1000\)) to 3.5 hours (\(\Gamma = 100\)).

\vspace{0.1cm}
\noindent \textbf{RQ 2:} \textit{How does the application of differential privacy affect the generation of mixed-type tabular data and the associated privacy risks in a federated learning setting?}

\vspace{0.1cm}

We assessed the impact of DP on fidelity, utility, and privacy risks across three privacy budgets \(\varepsilon \in [0.2, 1, 10]\) and without DP (\(\varepsilon = \infty\)) in a non-IID setting. The number of federated clients was fixed at \(\lambda = 5\) with client optimization rounds set to \(\Gamma = 100\). Privacy risks were measured by estimating Singling Out, Linkability, and Inference risks.

\vspace{0.1cm}

\begin{figure*}[t]
    \centering
    \begin{minipage}[b]{0.48\linewidth}
        \centering
        \includegraphics[width=\linewidth]{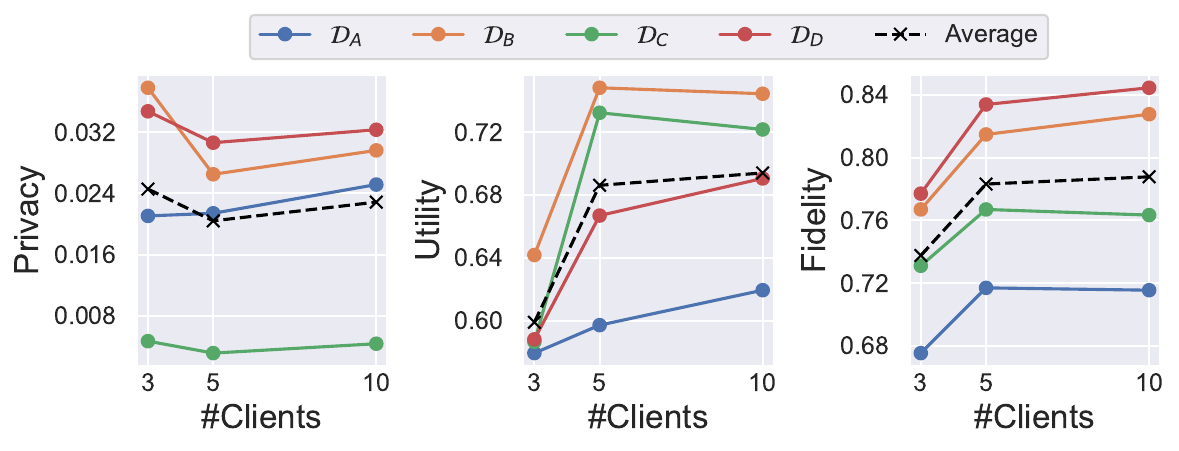}
        \caption{Privacy, utility, and fidelity evaluation using a varying number of federated clients \scalebox{0.95}{$\lambda \in [3,5,10]$} across all datasets ($\mathcal{D}_{A}$-$\mathcal{D}_{D}$) and non-IID setting. We kept the privacy budget \( \epsilon=1 \), local updates \( \Gamma=100 \) and FedAvg strategy fixed.}
        \label{fig:result_clients}
    \end{minipage}
    \hspace{0.02\linewidth}
    \begin{minipage}[b]{0.48\linewidth}
        \centering
        \includegraphics[width=\linewidth]{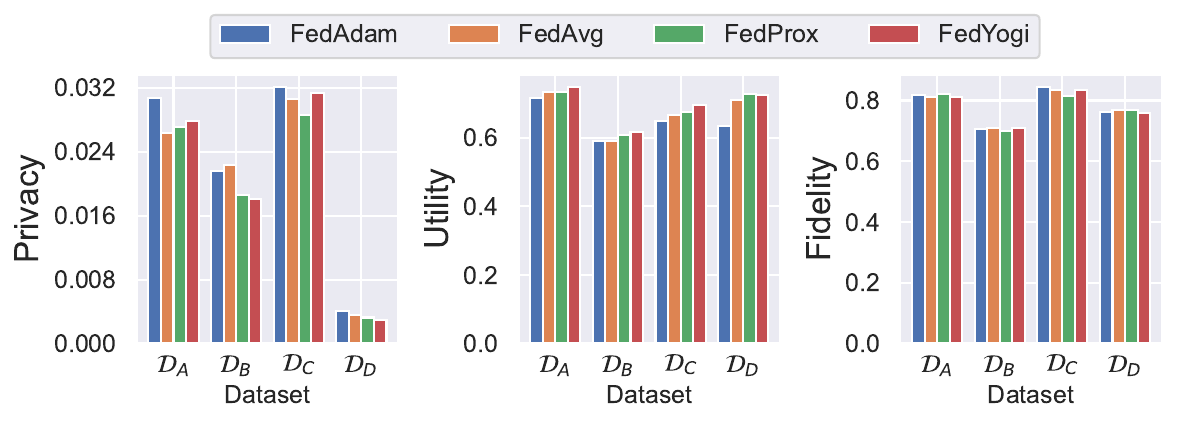}
        \caption{Privacy, utility, and fidelity evaluation using different federated strategies across all datasets ($\mathcal{D}_{A}$-$\mathcal{D}_{D}$) and non-IID setting. We kept the privacy budget \( \epsilon=1 \), local optimization updates \( \Gamma=100 \) and number of clients \( \lambda=5\) fixed.}
        \label{fig:results_strategies}
    \end{minipage}
\end{figure*}

\noindent \textbf{Results:} Smaller privacy budgets (lower \(\varepsilon\) values) enhance privacy protection but degrade data quality across all datasets (see \autoref{fig:DP_results}). Notably, a moderate privacy budget (\(\varepsilon = 1\)) increases privacy by~34\% while reducing utility by~15\% and fidelity by~14\% compared to the non-DP scenario (\(\varepsilon = \infty\)).

\vspace{0.1cm}

Our qualitative analyses, presented in \autoref{fig:TSNE}, support these observations. The 2D t-SNE representations demonstrate the preservation of sample relationships. Without DP (see \autoref{fig:tsne_noDP}), sample clusters closely resemble those in the original data (see \autoref{fig:tsne_real}). Introducing DP gradually alters the data structure, with more pronounced changes as privacy levels increase (see \autoref{fig:tsne_DP10} to \autoref{fig:tsne_DP02}). This progression reflects the trade-off between data utility and privacy enhancement.

\vspace{0.1cm}

Additionally, lower DP budgets (\(\varepsilon = 0.2\) and \(\varepsilon = 1.0\)) mitigate Singling Out, Linkability, and Inference risks, demonstrating the need for stringent DP constraints to minimize re-identification, linkage, and inference threats (see \autoref{fig:results_privacy_scores}). Higher DP budgets (\(\varepsilon = 10\) and \(\infty\)) reduce noise, enhancing data utility but compromising privacy. These results underscore the privacy-utility trade-off in synthetic data generation and the need for careful DP parameter tuning.

\vspace{0.1cm}

\noindent \textbf{RQ 3:} \textit{What is the impact of varying the number of federated clients and different strategies on the quality of generated mixed-type tabular data with differential privacy?}

\vspace{0.1cm}

\noindent We evaluated the impact on fidelity, utility, and privacy of synthetic data with three settings of federated clients \(\omega \in [3, 5, 10]\) using non-IID data partitions. The strategies compared included FedAvg, FedAdam, FedProx, and FedYogi. The number of client optimization rounds was fixed at \(\Gamma=100\) with a privacy budget of \(\varepsilon=1\).

\vspace{0.1cm}

\noindent \textbf{Results:} Increasing the number of federated clients enhances privacy protection, as it becomes more difficult to infer individual data points with more clients involved (see \autoref{fig:result_clients}). 

\vspace{0.1cm}

Additionally, fidelity and utility scores improve with a higher number of clients. We attribute this to the regularizing effect of DP, which reduces the client drift effect and results in more consistent local models. However, these benefits have limits. When expanding from five to ten clients, the gains in utility become marginal and can even decline. This performance degradation is likely due to the increased complexity of aggregating updates from a larger number of clients.

\vspace{0.1cm}

In terms of optimization strategies, there is no clear best choice when considering the trade-offs between fidelity, utility, and privacy. The strategies perform nearly identically in fidelity and utility, with slight decreases in performance due to unbalanced data distribution across clients (see \autoref{fig:results_strategies}). However, significant differences are observed in privacy performance across different datasets, indicating that dataset characteristics significantly influence the effectiveness of these federated learning strategies.

\vspace{0.1cm}

In summary, the evaluation of RQ 1, 2, and 3 reveals the trade-offs in federated learning with DP, highlighting the critical balance between privacy and data quality, as well as the influence of optimization strategies and federated configurations on the overall \textit{DP-FedTabDiff} model performance. 

\section{Conclusion}

We introduced \textit{DP-FedTabDiff}, a novel framework that unifies \textit{Differential Privacy} (DP), \textit{Federated Learning} (FL), and \textit{Denoising Diffusion Probabilistic Models} (DDPMs) to enable high-fidelity and privacy-preserving synthetic tabular data generation. By jointly addressing data utility and privacy, our framework offers a practical solution for secure data sharing and downstream analytics in sensitive, high-stakes domains such as finance or healthcare.

\vspace{0.1cm}

Our comprehensive evaluations revealed the trade-offs between data quality and privacy, showing that \textit{DP-FedTabDiff} maintains strong performance under realistic privacy budgets. These results highlight its robustness and potential as a foundation for privacy-aware generative modeling in distributed settings. Future work will explore adaptive strategies to dynamically balance privacy and data quality, further enhancing the applicability of federated learning in diverse settings.

\bibliographystyle{IEEEtran}
\bibliography{bibliography}

\end{document}